\documentclass[conference]{IEEEtran}

\IEEEoverridecommandlockouts

\usepackage{cite}
\usepackage{amsmath,amssymb,amsfonts}
\usepackage{algorithmic}
\usepackage{graphicx}
\usepackage{textcomp}
\usepackage{xcolor}
\usepackage{booktabs}
\usepackage{multirow}
\usepackage{fancyhdr} 

\def\BibTeX{{\rm B\kern-.05em{\sc i\kern-.025em b}\kern-.08em
T\kern-.1667em\lower.7ex\hbox{E}\kern-.125emX}}

\begin{document}

\title{Enhancing Clinical Decision-Making: Integrating Multi-Agent Systems with Ethical AI Governance}

\author{\IEEEauthorblockN{Ying-Jung Chen\IEEEauthorrefmark{1}\thanks{\IEEEauthorrefmark{1}YJC, AA conceive the idea, YJC, AA conduct research, YJC, AA, CSC write and edit the article}}
\IEEEauthorblockA{\textit{College of Computing} \\
\textit{Georgia Institute of Technology}\\
Atlanta, GA, USA \\
yjchen@gatech.edu}
\and
\IEEEauthorblockN{Ahmad Albarqawi}
\IEEEauthorblockA{\textit{MedWrite.ai} \\
Dublin, Ireland \\
\textit{University of Illinois} \\
Urbana, IL, USA \\
ahmada8@illinois.edu}
\and
\IEEEauthorblockN{Chi-Sheng Chen}
\IEEEauthorblockA{\textit{Neuro Industry Research} \\
\textit{Neuro Industry, Inc.}\\
Boston, MA, USA \\
michael@neuro-industry.com}
}

\maketitle

\thispagestyle{plain}
\fancypagestyle{plain}{
\fancyhf{} 
\fancyfoot[L]{979-8-3315-0266-9/25/\$31.00~\copyright2025~IEEE} 
\renewcommand{\headrulewidth}{0pt}
\renewcommand{\footrulewidth}{0pt}
}

\begin{abstract}
Recent advances in the data-driven medicine approach, which integrates ethically managed and explainable artificial intelligence into clinical decision support systems (CDSS), are critical to ensure reliable and effective patient care. This paper focuses on comparing novel agent system designs that use modular agents to analyze laboratory results, vital signs, and clinical context, and to predict and validate results. We implement our agent system with the eICU database, including running lab analysis, vitals-only interpreters, and contextual reasoners agents first, then sharing the memory into the integration agent, prediction agent, transparency agent, and a validation agent. Our results suggest that the multi-agent system (MAS) performed better than the single-agent system (SAS) with mortality prediction accuracy (59\%, 56\%) and the mean error for length of stay (LOS)(4.37 days, 5.82 days), respectively. However, the transparency score for the SAS (86.21) is slightly better than the transparency score for MAS (85.5). Finally, this study suggests that our agent-based framework not only improves process transparency and prediction accuracy but also strengthens trustworthy AI-assisted decision support in an intensive care setting.
\end{abstract}


\section{Introduction}
Artificial intelligence (AI) has been widely adopt into healthcare \cite{menzies2024ai}. Within medicine, it’s proving valuable for sharpening diagnostic precision, supporting treatment planning, and helping clinicians take care of patients \cite{li2023prediction}. Recent work has focused on using AI to interpret complex medical visuals like surgical footage \cite{lai2023intraoperative}, computed tomography (CT) scans \cite{deshpande2024combining}, and magnetic resonance imaging (MRI) scans \cite{dayarathna2024deep}, making interpretation faster and more consistent. These efforts provide  new possibilities across neurology, psychiatry, and continuous patient monitoring \cite{chen2024psycho}. Altogether, these advancements point to a future where AI supports both visual and signal-based insights, forming the backbone of smarter clinical decision-making tools.

Clinical decision support systems (CDSS) have become a vital part of current healthcare settings, offering insights drawn from electronic health records (EHRs) and real-time monitoring tools. Yet, many of the traditional AI models used in these systems fall short when it comes to flexibility, transparency, and oversight key qualities, especially critical in high-risk settings like intensive care units (ICUs). To address these limitations, we introduce a modular multi-agent system (MAS) designed to reflect how clinical teams make decisions, with a built-in emphasis on ethical AI to uphold both explainability and accountability.

Building on progress in large language model (LLM) agent-based frameworks, our system breaks down the decision-making pipeline into focused, collaborative agents. Each agent is responsible for a different aspect of ICU assessment: from interpreting lab results and tracking vital signs to making context-sensitive judgments based on a patient's history or co-existing conditions. These individual agents pass their findings to an integration agent that brings everything together, enabling more comprehensive predictions, examining its transparency, and cross-validating its outcomes. This system simulates how doctors gather evidence from various sources, weigh context, and form a unified clinical picture.

By structuring the system around modular agents and grounding it in ethical oversight, we improve not just how interpretable the model is, but also how it upholds accountability throughout the clinical decision-making process. To test the framework, we used the \textit{eICU Collaborative Research Database}~\cite{johnson2018eicu}, showing that our method can deliver well-organized predictions, shed light on key prognostic indicators, and build greater trust in AI-supported medical judgments.

\section{Related Work}
\subsection{Applications of Clinical Decision Support Systems in Intensive Care Settings}

 CDSS have come a long way, especially in ICU environments where every second counts. Earlier systems typically leaned on rule-based logic or statistical methods to generate recommendations \cite{arxiv240111120, ceurws2870}. More recent developments have looked to clinical practice guidelines (CPGs) as a way to enrich LLMs, boosting their ability to offer context-aware treatment advice. Research suggests that LLMs enhanced with CPGs outperform traditional models in delivering more accurate clinical suggestions \cite{arxiv240111120}.

Meanwhile, MAS to CDSS have been gaining popularity. Researchers developed a particularly interesting framework that combines case-based reasoning with a MAS. This system uses different agents to manage how users interact with it, execute tasks, and apply medical knowledge\cite{ceurws2870}. What makes this approach valuable is how it merges MAS with case-based reasoning, allowing the system to learn more efficiently and better adapt to each patient's unique situation.

\subsection{eICU Data and Its Applications}

The \textit{eICU Collaborative Research Database} has emerged as a critical resource for intensive care research, gathering comprehensive data from over 200,000 ICU stays across the United States \cite{nature2018eicu}. This extensive collection spans vital parameters—including vital signs, treatment protocols, severity indices, diagnoses, and interventions—serving as a solid groundwork for developing and validating AI models that address the specific challenges of critical care.

eICUs represent a significant influence on critical care medicine, harnessing telemedicine to address the shortage of on-site specialists for high-risk patients\cite{accesstelecareeicu}. This innovative method enables continuous expert monitoring and intervention without the limitation of physical distance. For example, Philips' eICU system uses the eCareManager platform to virtually bring ICU specialists to the patient's bedside. By connecting hospital networks and providing real-time clinical feedback, eICUs effectively bridge the gap between remote experts and immediate patient needs \cite{philipseicu}.

The implementation of eICU has been notable, as evidenced by the experience at Baptist Health South Florida. After introducing their eICU model, the institution saw a significant 23\% decrease in ICU mortality rates and up to a 25\% reduction in average length of stay (LOS)\cite{philipseicu}. These improvements really show how telemedicine is changing critical care for the better. The eICU approach makes care better in several important ways - doctors can watch patients around the clock, catch problems earlier, make better use of limited specialists, follow consistent treatment plans, and work more closely with the nurses and doctors at the bedside. Hospitals also receive financial benefits since patients are getting out of the ICU faster. In addition, all the data these systems collect is valuable for research, which can help to keep making ICU care better over time.

\subsection{LLM-Based Agents in Healthcare}

LLMs have recently become more common in healthcare. These models now assist in multiple areas, including virtual assistants, individualized health education, symptom checking, and mental health support tools \cite{linkedinllmagent}. By improving patient interactions and simplifying administrative tasks, LLM-based systems are beginning to influence how healthcare is provided.

One example is MDAgents, a MAS using LLMs to manage complex medical decisions \cite{arxiv240415155}. Its design replicates the teamwork observed in actual healthcare environments, enabling effective communication among its agents. Testing has shown that MDAgents performs better than earlier models in various evaluations.

A recent review explored the use of LLM-based agents in medical contexts \cite{arxiv250211211}. The review covered technical foundations, practical applications, and existing challenges. It emphasized components such as planning techniques, reasoning strategies, integration of external tools, and agent architecture. These systems are now employed for tasks like CDSS, automatic patient documentation, simulation training, and workflow optimization.

However, MedAgentBench, offers a benchmark for evaluating LLMs as medical agents, featuring 300 clinically-derived tasks across 10 categories and 100 realistic patient profiles. The results indicate that current LLMs still struggle with complex tasks, leading to the need for optimization prior to used in autonomous healthcare applications\cite{jiang2025medagentbench}.

Therefore, LLM-based agents have been further developed into MAS, where multiple agents interact in a collaborative manner. This change allows for systems that are more organized and flexible, offering new ways to manage challenging healthcare situations, such as emergency response coordination and personalized patient treatment.

\subsection{Multi-Agent Systems in Healthcare}

MAS are gaining attention as a promising way to tackle complex challenges in healthcare. One example applies MAS to pre-hospital emergency response, where agents—such as EMS dispatch centers, ambulances, traffic nodes, and medical providers—collaborate within a distributed decision-making setup \cite{pmc5547204}.

The idea of multi-stage AI agents builds on this by organizing intelligent agents into layers, each handling different parts of perception and reasoning. Many of these layers are now powered by LLMs, allowing for more structured and scalable workflows \cite{linkedinmultistage}. This layered setup has shown promise in areas like personalized care and remote health services.
Furthermore, the LLM-medical-agent framework, for instance, demonstrates how MAS can be applied to modular analysis of healthcare data in practical settings \cite{llmmedicalagent}.

\subsection{Ethical Governance in Healthcare AI}

Ethical rules applied in healthcare AI is addressed through Explainable AI (XAI), which showcases the decision-making processes of complex algorithms. "Healthcare AI Datasheets" framework, which  documents potential biases through demographic data\cite{siddik2025datasheets}. These complementary approaches not only enhance transparency by demystifying the "black-box" problem but also actively work toward equitable healthcare outcomes by identifying and addressing the sampling and complexity biases that have historically perpetuated healthcare disparities across diverse populations\cite{clearstepxai}.

Growing concerns around the safety of LLM-based agents have prompted the development of MAS \cite{chen2024information,radanliev2025ai}, which embed ethical advisor and policy guardrails to ensure compliance with safety and privacy standards—an especially important safeguard in clinical environments \cite{guardagentpaper}.

Bringing LLMs into electronic health record systems also introduces a range of ethical, legal, and practical questions. These include how to handle consent, maintain oversight, and ensure data governance \cite{natureehr2025}. A patient-centered approach—with transparency and strong ethical foundations—is essential for protecting vulnerable groups.

Thus, the World Health Organization (WHO) has outlined ethical guidelines for AI in healthcare, highlighting principles such as human autonomy, well-being, and system transparency \cite{whogov}. Especially in high-risk areas like intensive care units, solving these governance challenges is key to the responsible deployment of AI \cite{scirp2024icu}.

\subsection{Motivation and Research Gap}

While AI-powered CDSS have been promisingly adopted, there are still needs to be considered for real-world ICU practices. Many solutions fall short in modularity due to lack of transparency, or aren't built with an inter-agent communication system to reflect the dynamic, interdisciplinary nature of intensive care.

Currently, most existing approaches tend to focus on isolated tasks—like interpreting lab results, monitoring vital signs, or reasoning based on medical history—but few bring these components together into a unified or dynamic system that reflects how clinicians actually work as a team. The essential needs draw our attention to propose a novel agent-based system that breaks down the clinical reasoning process.  

Thus, we build the system with specialized, collaborative agents—each designed to handle a distinct aspect of care while maintaining accountability and interpretability throughout. By applying ethical AI principles at every stage of the pipeline and validating our design using the eICU database, this study aims to bridge both the technical and ethical gaps in deploying trustworthy AI for high-stakes decision-making in critical care.

\begin{figure*}
    \centering
    \includegraphics[width=.8\linewidth]{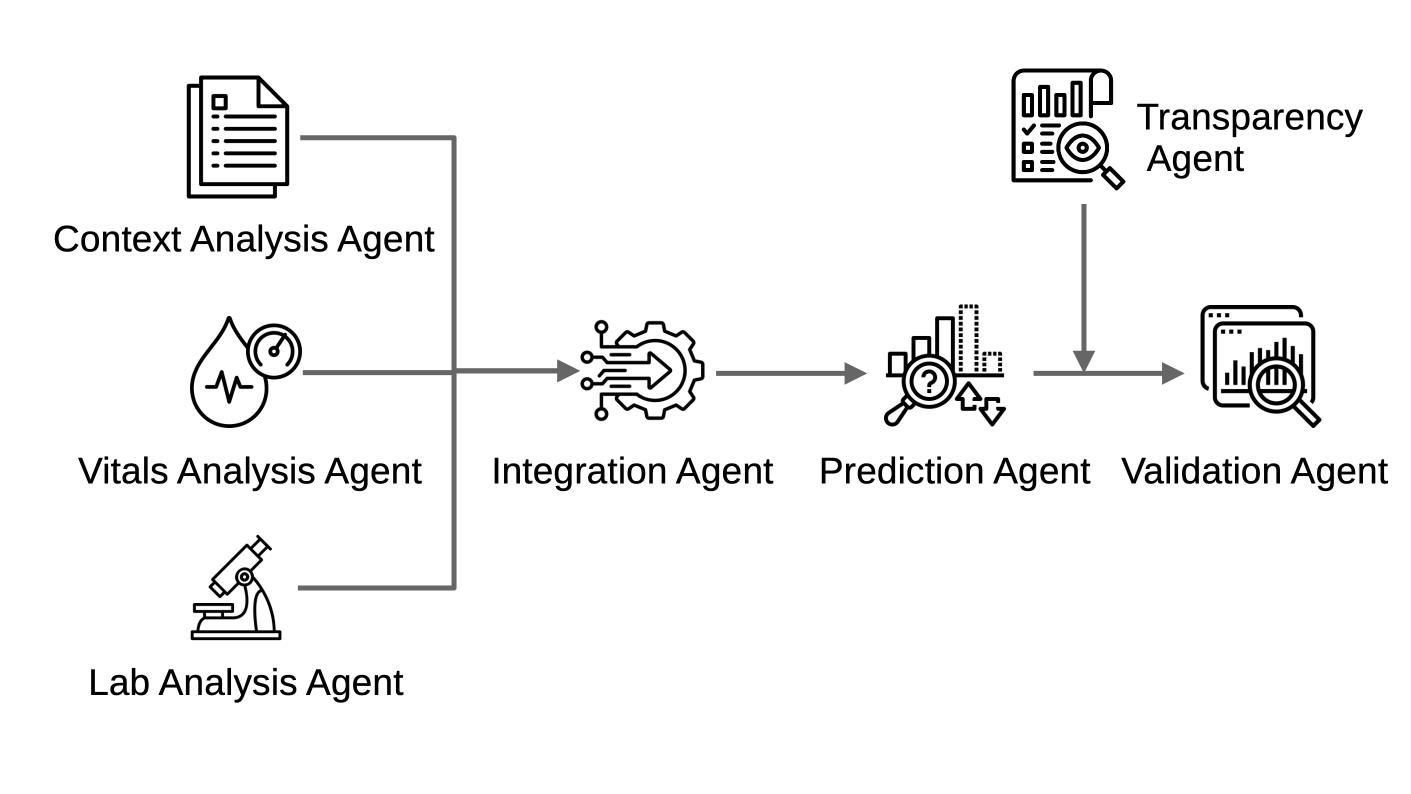}
    \caption{Illustration of the MAS Design.
The system consists of a set of specialized agents, each responsible for processing a specific type of clinical data. The Context Analysis Agent handles unstructured inputs like clinical notes, while the Vitals Analysis Agent focuses on real-time physiological signals, and the Lab Analysis Agent interprets laboratory test results. These distinct streams of information are brought together by the Integration Agent, which fuses multimodal features into a unified representation. Based on this, the Prediction Agent carries out key forecasting tasks—such as predicting ICU mortality or estimating LOS. To support interpretability, the Transparency Agent generates human-readable, traceable explanations of model outputs. Finally, the Validation Agent oversees performance assessment by comparing predictions against ground truth data.}
    \label{fig:gabs}
\end{figure*}

\section{Methods}
\subsection{Dataset and Preprocessing}

This study used the \textit{eICU Collaborative Research Database v2.0}~\cite{nature2018eicu}, which compiles anonymized ICU records from over 200,000 patient admissions in different hospitals across the U.S. The database contains two types of data: 1) structured details (e.g., vital signs and lab results) and 2) unstructured clinical notes contributed by nurses and physicians, giving us a overall view of patient care.

Here we used several key eICU files: \texttt{patient.csv}, \texttt{lab.csv}, \texttt{vitalPeriodic.csv}, \texttt{note.csv}, and \texttt{medication.csv}, along with APACHE-related data files \cite{knaus1981apache} (\texttt{apacheApsVar.csv} and \texttt{apachePatientResult.csv}). To align each patient's information, records were grouped according to \texttt{patient-unit-stayid}. If any essential data were missing—such as vital signs, lab values, or clinical notes—we removed those entries to maintain reliability.

We filled in the data gaps, ordered events based on their timestamps, and shortened lengthy text fields to meet the language model guidelines. Then, we sampled 150 patients for the study: 76 mortality patients and 74 survived patients. This balanced sample, provided a controlled dataset for our comparative analysis.

Then, we retrieved specific features from each patient's record set. We collected the ten most recent vital sign readings to reveal each patient's current physiological state. We also selected the latest distinct lab biomarkers deemed clinically relevant. When dealing with unstructured clinical documentation, we included up to three notes for every patient, focusing primarily on entries written by physicians and nurses. In addition, our analysis tracked the top 20 medications and treatments, identifying them by frequency or uniqueness within the overall dataset. Finally, we incorporated APACHE scores and predictions as reference points to aid in validating and evaluating our modeling outcomes.

\subsection{Multi-Agent System Design}

To emulate real-world ICU decision-making, we implemented a modular MAS consisting of six discrete agents, each responsible for a semantically distinct task. The system is shown in Figure~\ref{fig:gabs}.

\begin{itemize}
  \item \textbf{Lab Analysis Agent:} Receives structured lab data and highlights key abnormalities (e.g., hyperlactatemia, creatinine elevation) with implications on APACHE scoring and patient prognosis.
  \item \textbf{Vitals Analysis Agent:} Processes vital signs (e.g., heart rate (HR), systolic blood pressure (SBP), peripheral capillary oxygen saturation ($SpO_{2}$), temperature) and evaluates physiological stability, respiratory function, and cardiovascular performance.
  \item \textbf{Context Analysis Agent:} Analyzes unstructured notes, medication usage, and treatment strings to infer diagnoses, risk factors, and progression trajectory.
  \item \textbf{Integration Agent:} Aggregates all the results from the above agent into a comprehensive, system-by-system clinical assessment. It prioritizes ICU risk factors related to mortality and length of stay (LOS).
  \item \textbf{Prediction Agent:} Generates structured outcome predictions (mortality probability and ICU LOS) using integrated findings and APACHE variables. Results follow a strict template for automated parsing.
  \item \textbf{Transparency Agent:} Analyzes clinical prediction outputs to meet ethical standards and are explainable to various stakeholders. This score includes calculating explainability, interpretability, and traceability scores based on specific steps for each dimension. 
  \item \textbf{Validation Agent:} Compares predicted vs. actual ICU outcomes and reflects on the prediction’s accuracy, key contributing variables, and future improvement insights.
\end{itemize}

To ensure information transfer between agents, we implemented a shared memory architecture that allows any agent to access inputs and outputs from previous pipeline stages. This approach reduces the risk of information loss between modules while maintaining the semantic separation of responsibilities. While this shared memory design has proven effective at improving predictive performance by ensuring that no critical data leakage during the analysis process, we recognize that the MAS introduces additional complexity that can impact transparency. Our current implementation focuses on performance optimization, with ongoing work to enhance the clarity across the agent communication pathways.

Each agent is implemented using OpenAI GPT-4o \cite{openai2024gpt4o} and configured via the intelli framework \cite{intelli2024}, which allows asynchronous agent orchestration with JSON-structured prompts and logging.

\subsection{Few-Shot Learning Example Construction}

To ground the reasoning of the Prediction Agent, we incorporated two real ICU patients as few-shot exemplars. These examples span different outcomes (e.g., survived vs. expired) and are selected based on APACHE completeness and data richness. Each example includes demographics, APACHE variables, labs, vitals, and actual outcomes. The examples are embedded directly in the prompt using clearly segmented format blocks and used to improve model generalizability.

\subsection{System Execution and DAG Orchestration}

The entire multi-agent pipeline is expressed as a directed acyclic graph (DAG), where tasks are mapped via semantic dependencies. Specifically:
\begin{itemize}
  \item \texttt{lab\_analysis}, \texttt{vitals\_analysis}, and \texttt{context\_analysis} feed into \texttt{integration}.
  \item \texttt{integration} feeds into \texttt{prediction}.
  \item \texttt{prediction} feeds into \texttt{transparency}.
  \item \texttt{transparency} feeds into \texttt{validation}.
\end{itemize}

Execution is managed asynchronously using Python’s \texttt{asyncio} to allow concurrent LLM calls and reduce latency. The system supports multi-threaded batch evaluation and error-tolerant retries.

\begin{figure*}
    \centering
    \includegraphics[width=1\linewidth]{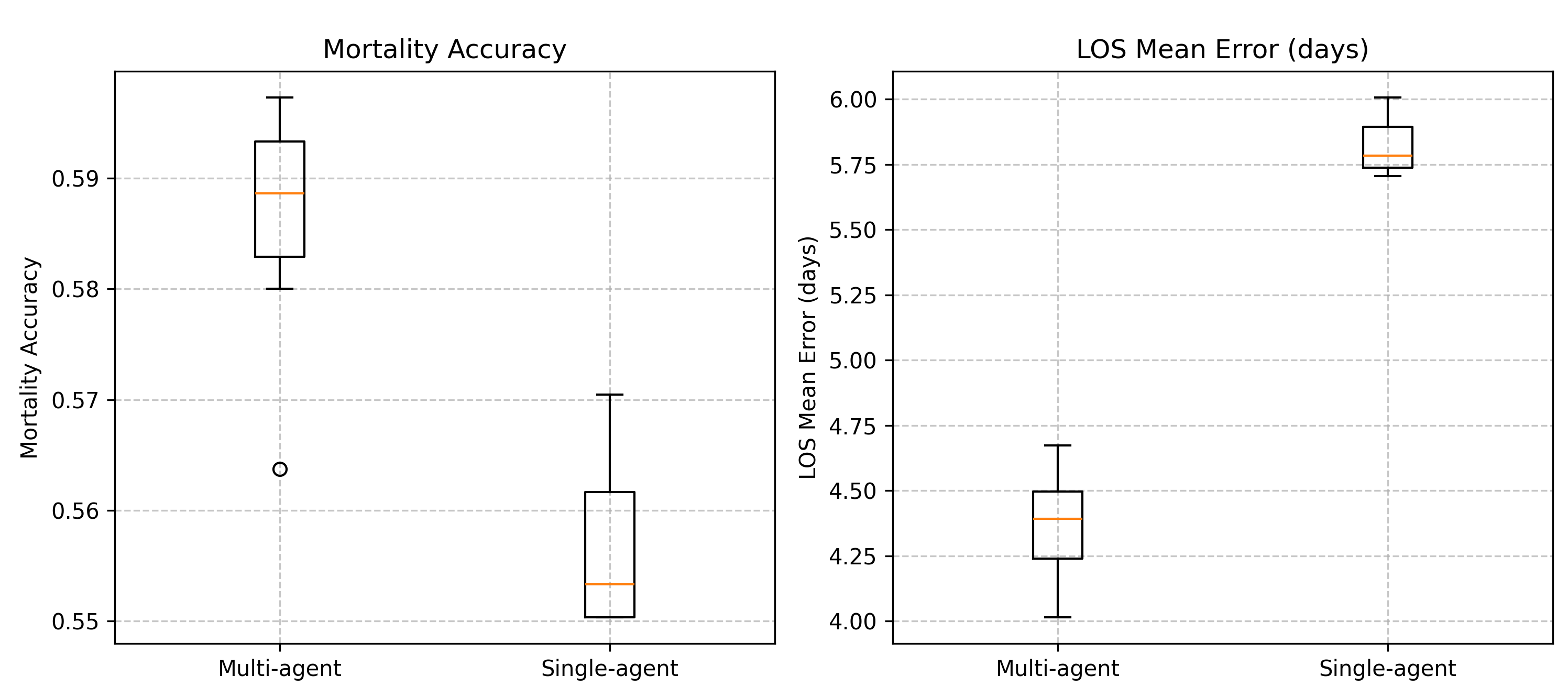}
    \caption{Performance comparison between the MAS and SAS across two evaluation metrics: \textbf{Mortality Prediction Accuracy} and \textbf{Length of Stay (LOS) Mean Error}. Each model was executed 8 times, and the box plots represent the distribution of results across these runs. The MAS demonstrates slightly higher mean mortality accuracy and consistently lower LOS errors than the SAS.
    }
    \label{fig:res}
\end{figure*}

\subsection{Implementation Details}

Each agent is instantiated as a generative pre-trained transformer-based (GPT-based) \cite{radford2018improving} text agent with a predefined \texttt{mission}, API credentials, and output format enforcement. Prompts are customized per task using structured sections (e.g., “KEY ABNORMALITIES”, “APACHE RELEVANT FINDINGS”). Inputs are truncated or summarized to fit within the 10,000-token limit of GPT.

Agent configuration example:
\begin{verbatim}
Agent(
  provider="openai",
  mission="Analyze lab data for 
  abnormalities",
  model_params={"key": OPENAI_API_KEY, 
  "model": "gpt-4o"}
)
\end{verbatim}

All patient data is saved per analysis run, including intermediate and final agent outputs in JSON format.

\subsection{Ethical AI and Explainability}

Here we incorporate the explainability,interpretability, and traceability aspects within our agent system:
\begin{itemize}
  \item \textbf{Explainability:} Measures how the system detects and explains the factors influencing predictions. This includes elements such as the patient's age, intubation status, creatinine levels, and other clinical parameters.
  \item \textbf{Interpretability:} Evaluates how the reasoning process is communicated among various stakeholders. This step can allow the decision-making logic is understandable to clinicians, patients, and administrators.
  \item \textbf{Traceability:} Assesses the trace process starting from input data, as well as influencing factors back to their sources. This builds a solid trail of evidence backing up the system's findings, making it easier to trust and use in real-world scenarios.
\end{itemize}

\subsection{Quantitative Evaluation of Agent Transparency}
We built the transparency assessment process, which is similar to \cite{CAWIDING2025109589} evaluating the transparency of clinical prediction explanations by analyzing text responses for key features. The transparency score is calculated based on explainability, interpretability, and traceability in the clinical prediction framework. The explainability score shows how the system justifies decisions based on feature importance identification, concentration of critical factors, clear reasoning, and stakeholder explanations. In addition, the interpretability measures how humans understand the model's workings based on reasoning processes, prediction predictability, complexity, and alternative scenario analysis. The traceability assesses documentation quality by tracking input data sources, data transformations, model development history, and decision process. The overall transparency score is calculated by these components, showing strong performance but it indicates needs for improving interpretability.

\section{Results}

Our results shown in Table~\ref{tab:res} and Figure~\ref{fig:res} demonstrate that the MAS consistently outperforms the SAS approach across all evaluated metrics over 150 unique patients. Each experiment was conducted across eight runs with approximately 150 patients per run, and the reported values represent the average performance to ensure consistency and robustness of evaluation. In terms of mortality prediction accuracy, the MAS achieved a mean of 59\%, while the SAS reached only 56\%. While this 3 percentage point difference was consistent across multiple runs, further statistical analysis with larger samples would be necessary to establish clinical significance. Additionally, the standard deviation for the MAS is marginally higher, suggesting a bit more variability across runs.

We conducted paired t-tests across all performance metrics. The results show that these improvements are not just numerical differences but statistically robust findings, with p-values well below 0.0001 for most metrics and p = 0.0001 for mortality accuracy. What's particularly encouraging is that the confidence intervals for all improvements exclude zero, meaning we can be confident these gains are real rather than chance variations. This statistical validation directly addresses a key concern in clinical AI research - distinguishing between genuine performance improvements and random noise in the data.

\begin{table}[h]
\centering
\caption{Statistical Comparison of Multi-agent and Single-agent Models (Average over 8 runs)}
\small
\begin{tabular}{lccc}
\toprule
\textbf{Metric} & \textbf{Model} & \textbf{Mean (SD)} & \textbf{p-value} \\
\midrule
\multirow{2}{*}{\begin{tabular}{@{}c@{}}Mortality Prediction\\Accuracy (\%)\end{tabular}} 
    & Multi-agent   & \textbf{58.6 (1.1)}  & \multirow{2}{*}{\textbf{0.0001}} \\
    & Single-agent  & 55.7 (0.8)  & \\
\midrule
\multirow{2}{*}{\begin{tabular}{@{}c@{}}LOS Mean\\Error (days)\end{tabular}} 
    & Multi-agent   & \textbf{4.37 (0.21)}  & \multirow{2}{*}{$\mathbf{p < 0.0001}$} \\
    & Single-agent  & 5.82 (0.11)  & \\ 
\midrule
\multirow{2}{*}{\begin{tabular}{@{}c@{}}Mean Squared\\Error (days²)\end{tabular}} 
    & Multi-agent   & \textbf{35.5 (2.3)} & \multirow{2}{*}{$\mathbf{p < 0.0001}$} \\
    & Single-agent  & 48.1 (1.4)  & \\
\midrule
\multirow{2}{*}{\begin{tabular}{@{}c@{}}Root Mean Squared\\Error (days)\end{tabular}} 
    & Multi-agent   & \textbf{5.95 (0.19)} & \multirow{2}{*}{$\mathbf{p < 0.0001}$} \\
    & Single-agent  & 6.94 (0.10)  & \\
\bottomrule
\end{tabular}
\label{tab:res}
\end{table}

Our analysis indicates an enhancement in predicting LOS when using the MAS. In our study, the average prediction error drops to 4.37 days under the MAS, compared to 5.82 days observed with the SAS—an improvement of roughly 25\% in accuracy. This gain is important, given its direct influence on how ICU resources are allocated and care is planned. In addition, the MAS registers a mean squared error of 35.49 as opposed to 48.13, along with a root mean squared error of 5.95 compared to 6.94, signifying that it delivers more stable and consistent predictions with fewer extreme fluctuations.

A particular aspect of our findings is how the MAS manages to reduce the LOS prediction error. A mean error of 4.37 days, as opposed to 5.82 days from the SAS, illustrates a noteworthy improvement of 25\%, which is critical in fine-tuning patient care. Moreover, the improved metrics—lower mean squared error (35.49 instead of 48.13) and root mean squared error (5.95 compared to 6.94)—further confirm that this system not only enhances accuracy but also offers more stable predictions across various patient groups.

\begin{figure}[htb!]
    \centering
    \includegraphics[width=1\linewidth]{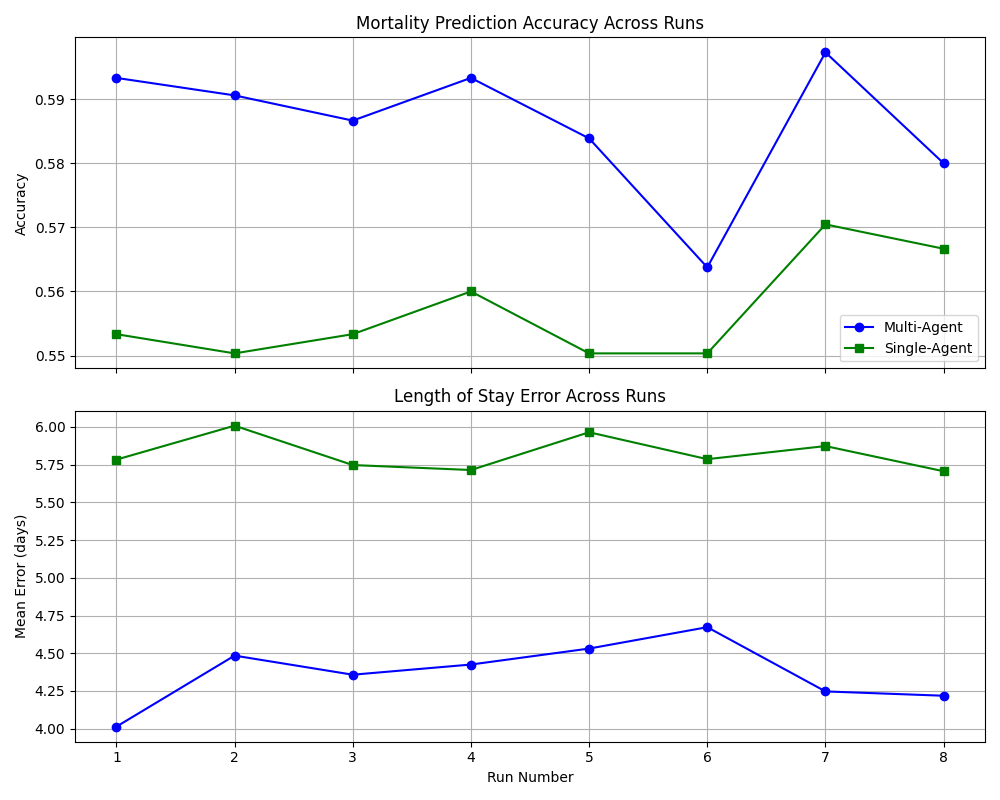}
    \caption{A comparison of MAS and SAS performance across two key metrics: \textbf{Mortality Prediction Accuracy} (top) and \textbf{Length of Stay Mean Error} (bottom). The diagram displays results from eight independent runs. For mortality prediction, higher accuracy indicates better performance, while lower mean error values (in days) represent more accurate predictions for LOS.}
    \label{fig:run_comparison}
\end{figure}

Figure~\ref{fig:run_comparison} reveals a clear analysis across the eight test runs we conducted. Looking at the top graph, you can see the MAS (blue line) consistently outperformed the SAS (green line) in predicting mortality. The MAS accuracy ranges from about 56\% to nearly 60\%, while the SAS stays between 55\%--57\%. What's notable not just that it performed better, but that this advantage held steady across every single run. 

The LOS prediction results in the bottom graph show more improvement for MAS. The blue line stays below the green throughout all runs, with errors around 4--4.7 days compared to the Single-agent's 5.7--6 days. That gap - somewhere around a day and a half - might not sound huge until you consider what it means for real patients and hospital planning. Most systems like this show more ups and downs, but here the MAS maintained its edge consistently. 

\begin{table}[h]
\centering
\caption{Comparison of transparency score of MAS and SAS (Average over 8 runs)}
\begin{tabular}{lcccc}
\toprule
\textbf{Metric} & \textbf{Model} & \textbf{Mean} \\
\midrule
\multirow{2}{*}{Average transparency score (\%)} 
    & Multi-agent   & 85.50  \\
    & Single-agent  & \textbf{86.21}  \\
\bottomrule
\end{tabular}
\label{tab:res2}
\end{table}

Table~\ref{tab:res2} provides a side-by-side comparison of average transparency scores from eight independent runs. The data shows that the SAS approach scores an average of 86.21\% in contrast to 85.50\% for the MAS. This suggests that, under our current evaluation criteria, both models perform similarly in terms of transparency with the SAS design being marginally more interpretable. Despite the distributed nature of the MAS, it maintains nearly equivalent transparency levels, indicating that our shared memory architecture effectively preserves reasoning traceability across multiple specialized agents.

Overall, the MAS outperforms SAS in terms of predictive accuracy and error. Interestingly, our results provide a different perspective from previous research \cite{cemri2025multiagent} that suggests MAS may fail through inter-agent misalignment. The shared memory architecture within MAS and transparency assessment process can establish a trustworthy CDSS. This finding highlights that a well-optimized inter-agent dynamic is more beneficial than a simpler SAS. Although our MAS offers a robust system, we still need to consider whether additional factors—such as scalability or deployment constraints—demand a MAS, based on the results observed.

\section{Conclusion}

In this study, we found that the MAS performs better than the SAS for the mortality prediction accuracy (59\% vs. 56\%, respectively). 
Additionally, our results indicate that MAS has a shorter predicted LOS while maintaining a similar degree of transparency. Specifically, the MAS scores 85.5\% for transparency, whereas the SAS shows a very close 86.21\%. This subtle difference suggests that our MAS effectively maintains transparency despite the nature of complicated coordinating decisions among several specialized agents.

The balance between performance and explainability in decision support systems has significant implications for clinical patient care. In critical care scenarios where prediction accuracy is important, our MAS approach presents significant advantages. While maintaining high transparency, the MAS system improves predictive performance and addresses the critical needs for both reliable outcomes and interpretable decision processes that healthcare professionals can incorporate into their clinical judgment. On the other hand, when explaining prediction rationales to build clinician trust is essential, the higher transparency within Single-Agent Systems (SAS) still needs further investigation. Our research will focus on enhancing the clarity of the MAS while preserving its predictive advantages. Through improved inter-agent communication protocols and more sophisticated explanation mechanisms, we aim to develop a system that combines the predictive capabilities of MAS with the interpretability essential for safe and effective implementation in critical care settings.

\section*{Acknowledgment}

The authors thank for Medwrite Limited for support our work.

\bibliographystyle{IEEEtran}

\bibliography{ref}

\end{document}